\documentclass{llncs}
\usepackage{llncsdoc}
\usepackage{geometry}   
\usepackage[center]{caption}
\usepackage{subfigure}
\usepackage{qtree}  
\usepackage{tikz}            % See geometry.pdf to learn the layout options. There are lots.

\usepackage{graphicx}
\usepackage{amssymb}
\usepackage{amsmath}
\usepackage{array}
\usepackage{mathtools}
\mathtoolsset{showonlyrefs}

\newcounter{rowcount}
      \newtheorem{thesis}{Thesis}
\title{Deontic Logic for Human Reasoning}
\author{Ulrich Furbach and Claudia Schon\thanks{Work supported by DFG FU 263/15-1 'Ratiolog'}        \\\small{\{uli,schon\}@uni-koblenz.de}}
\date{} 
\institute{Universit\"at Koblenz-Landau}
\begin{document}
\maketitle
%\section{}
%\subsection{}

\begin{abstract}  
Deontic logic is shown to be applicable for modelling human reasoning. For this the Wason selection task and the suppression task are discussed in detail. Different versions of modelling norms with deontic logic are introduced and in the case of the Wason selection task it is demonstrated how differences in the performance of humans in the abstract and in the social contract case can be explained. Furthermore it is shown that an automated theorem prover can be used as a reasoning tool for  deontic logic.
\end{abstract}
\vskip 3ex
\section{Introduction}
Human reasoning and in particular conditional reasoning has been  researched in various disciplines. In cognitive psychology a lot of experimental data is collected and there are numerous  different modelling approaches. In philosophy, rationality and  normative reasoning is a topic with increasing interest. In artificial intelligence research the aim is to model human rational reasoning within artificial systems. 

Recently there are some papers from automated reasoning  which try to model experiments from cognitive psychology; in particular the experiments involving the Wason selection and the suppression tasks are discussed in the literature (\cite{holldobler2009logic,holldobler2011abductive}).

In this paper we want to contribute to this discussion by advocating deontic logic to this end. We are well aware that this is not the first paper proposing deontic logic for conditional reasoning. However, our aim is not only to use this  logic to model the settings and the result of these experiments, moreover, we want to use an \emph{automated reasoning system} to solve the tasks. 
There are only few automated theorem provers specially dedicated for deontic logic and used by deontic logicians (see \cite{Artosi94ked:a,Bassiliades:2011:MDR:2441484.2441486}). Nonetheless, several approaches to translate modal logic into (decidable fragments of) first-order predicate logics are stated in the literature. A nice overview including many relevant references is given in \cite{SH13}. 
We will use the first order predicate logic prover Hyper for deontic logic, which is possible because we translate the latter into the description logic  $\mathcal{ALC}$. This again can be translated into DL-clauses, for which Hyper is a decision procedure.

\begin{figure}[t]
\centering
\subfigure[ If there is a vowel on one side, the opposite side contains an even number. ]{\label{wason-a} 
  \includegraphics[height=.1750\linewidth]{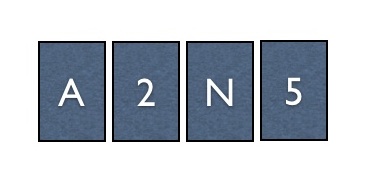}
}\hspace{0.1\linewidth}
\subfigure[If a person is less than 21 years old, she is not allowed to drink beer.]{\label{wason-b}\includegraphics[height=.175\linewidth]{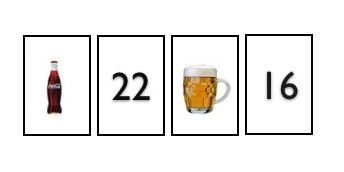}
}
\caption{The Wason Selection Task}
\label{fig:wason}
\end{figure}

%\section{The Wason Selection Task}
The Wason selection task  (WST) was first presented by the psychologist Peter C Wason in \cite{wason1968reasoning} and is one of the most carefully researched experiments in the area of human rational reasoning. The abstract case of the task is shown in part (a) of Fig.~\ref{fig:wason}. In the task, four different cards are presented to a test person. The test person is told, that each card contains a letter on one side and a number on the opposite side. Further a statement like  ``If there is a vowel on one side of the card the opposite side contains an even number'' is given. Now the test person is asked to verify/falsify this statement by turning a minimum number of cards. In this abstract task, less then 25 \% of the test persons were able to find the solution.
%In \cite{stenning2008human} the WST is discussed in deep detail. There also is a  collection of different approaches using various logics for modelling the selection task in a way that the results from the experiments are captured. 
%\remarkx{D\"urfen wir die Bilder von dem Bier und der Cola verwenden?}
It is most obvious (at least for a logician) to understand the conditional which formulates the task as a material implication. That means that the implication $P \rightarrow Q$ can be replaced by the formula $ \neg P \lor Q$. Assume that property $P$ holds, then, for $ \neg P \lor Q$ to be true, $Q$ has to be true. Very many experiments have shown  that humans have problems to perform this inference properly. If context is added to the problem people solve the problem with a much higher correctness rate.
By adding additional context, the problem can be a social contract problem or a precaution problem. 
One example for a social contract context as addressed in part (b) of Figure~\ref{fig:wason}, is a setting, in which one side of the cards shows a beverage, namely beer or lemonade and the other side the age of the person who drinks this beverage. The rule is ``If a person is less than 21 years old she is not allowed to drink beer''.  In this case 75 \% of the subjects gave the correct solution.

A different class of contexts can be formulated by a so-called ``precaution rule'', e.g. rules of the form ``If you agree in a hazardous activity then you must take the precaution''.  In this case, like in the social contract context, people perform dramatically better, compared to the abstract case. 

In the following section we discuss several logical approaches to model the WST. In Section~\ref{sec:deontic} we introduce our approach using deontic logic. Section~\ref{sec:suppression} models the suppression task and in Section~\ref{sec:consistency} we show how our approach can be used to check the consistency of normative systems automatically. For a conclusion we briefly comment on attempts to formulate a kind of 'robot ethics'.
%\remarkx{Ist die \"Ubersicht ausf\"uhrlich genug?}

\section{Logical Models for the WST}
\label{sec:models}
Since the WST is dealing with conditional reasoning it seems to be  natural to use predicate logic for modeling the task and to use existing logical inference mechanisms  to model human reasoning. A very careful discussion of various logics to this end can be found in \cite{stenning2008human}; in particular all these investigation into logics are nicely combined with findings about the psychology of human reasoning. One of these logics,  multi-valued logic, seems to be very likely in the case of the WST, where the invisible side of a card can easily be modeled by the truth value ``unknown''. Several authors apply multi-valued logic to model human reasoning, e.g., \cite{holldobler2009logic} use a Lukaswiewicz logic together with logic programs. In \cite{holldobler2011abductive} this approach is combined with the concept of abduction, which is proposed in \cite{kakas1992abductive} and it is also used for modeling human reasoning in \cite{kowalski2011computational}.
%\remarkx{Kowalski Buch muss noch gecheckt werden.}

All these approaches use logic programming for modeling human reasoning. However, one should have in mind, that logic programming  languages and its semantics have been designed for \emph{programming}. There are at least three main issues of logic programs as used in the cited approaches:
\begin{itemize}
\item The language is restricted to definite clauses, i.e. clauses of the form $A \leftarrow B_1\land \cdots \land B_n$, where the left-hand side, the head,  contains only one atom, and the right-hand side, the body, contains a conjunction of literals. This special form does not allow the representation of a disjunction, say $A \lor B$. This is not a problem for programming purposes, because one can easily  show that every Turing computable function can be represented by definite clause programs. However, for the modeling of human reasoning it should be possible to express  disjunctions.\footnote{In Artificial Intelligence we very well remember the relapse in the development of artificial neural networks, when the observation that perceptrons cannot compute a disjunctive or was spread.} 
\item The right hand side of a clause can contain literals, i.e. the negation of atoms. This negation, however, is not a logical negation, it is a non-monotonic negation, which usually is based on a closed-world assumption.
\item The semantics of logic programs with non-monotonic negation involves either so-called completion mechanisms or interesting fixed-point operations to construct models. For all of these model construction mechanisms it turns out that they involve much more complex reasoning, compared to the monoton case. Furthermore we doubt that those constructions are easily accessible to humans and their inference mechanisms.
\end{itemize}
The extension of logic programs with abduction turns the clauses, the logical rules, into  licences for conditionals using abnormality predicates: $A  \leftarrow B_1\land \cdots \land B_n\land \lnot ab$,
with the reading ``If nothing abnormal is known and all the $B_i$ hold, then $A$ holds''. Note that the negation symbol in front of the $ab$ atom is the  non-monotonic negation as mentioned before. We propose to model this distinction of normal from abnormal  behaviour by introducing an explicit operator instead of coding it into the clauses; just use deontic logic.

\section{Deontic Logic and the WST}
\label{sec:deontic}
The difference in behavior between the abstract case of the WST, the social contract and the precaution problem leads immediately to a distinction between descriptive and deontic conditionals. A deontic interpretation of the rules from the WST leads to a description of a  \emph{norm}; hence the rule makes a statement about how the world \emph{ought to be}.

There is an ongoing discussion about the use of deontic logic. In \cite{stenning2008human} the authors explicitly discuss deontic logic as a modal propositional logic for the WST. They construct models for a specification  of the selection task, but they do not discuss the representation of the task itself in deontic logic. Another detailed investigation of deontic logic can be found in \cite{cosmides2008can}, where the authors give an overview from a psychological and neurobiological point of view. They further discuss the deontic nature of the selection task in various contexts. There is the purely declarative version, which corresponds in our example to the vowel--consonant version and a social contract version, e.g., the beer--age version. Cosmides et al further argue that there is also  the class of the precaution rules as  introduced above. The different nature of these  contexts causes the authors in \cite{cosmides2008can} to conclude that there cannot be a general deontic logic for capturing human reasoning about conditionals. Indeed, there is strong evidence that humans have different reasoning mechanisms available depending on the nature of the reasoning task. There is the case of a patient, R.M.,  reported in \cite{stone2002selective}, who had a severe accident and suffered from  severe  retrograde amnesia. The damage of his brain was  in different areas of the cortex, such that both sides of the amylgada were disconnected. The authors made extensive reasoning  experiments with R.M. using 65 reasoning tasks based on the WST. It turned out that R.M.'s performance on the abstract reasoning problems (16,7 \%) and on the precaution rules was comparable to controls (70 \%), whereas the score on social contract problems was 31 percent points lower. This clearly indicates that there are different reasoning mechanisms for those  contexts.
In \cite{cosmides2008can} the conclusion from these findings is that there is no general deontic logic applicable for the modeling of this behavior. We support this hypothesis and at the end of Section \ref{sect:abstract vs context WST} we discuss a multi modal logic which very well is able to model these diverse kinds of reasoning.

Another observation discussed in \cite{cosmides2008can} is, that the WST in general can be seen as a cheat detection task. In different social contexts humans may apply different inference systems for  cheat detection. %We will argue, that this cheating detection question has to be modeled by mechanisms different from the usual question of logic consequence, which is employed more or less explicitly by automated deduction systems.

In the following deontic logic as a modal logic is introduced and used to formalize the WST. %Further we show how to use an automated theorem prover to solve the WST.

\subsection{Deontic Logic as Modal KD}

Deontic logic is a well studied modal logic very suitable to model human reasoning. It corresponds to the modal logic K together with a seriality axiom D:
\[  \textrm{D:}\hspace{2em}    \Box \Phi \rightarrow \Diamond \Phi \]
In contrast to K, the $\Box$-operator is interpreted  as `it is obligatory that' and the $\Diamond$ as `it is permitted that'.

In modal logic, semantics are given by so called Kripke structures consisting of a set of \emph{possible worlds} connected by a reachability function. Each world is labeled by the set of formulae, which are true in the respective world. A formula of the form $\Box F$ is read as ``in every reachable world, $F$ is true''. 
Hence if $w$ is a world we have
\begin{equation*}
w \models \Box \Phi  \hspace{2em} \textrm{iff} \hspace{2em} \forall v : R(w,v) \rightarrow v \models \Phi
\end{equation*}
A formula $F$ is called satisfiable, if there is a Kripke structure and a world in which $F$ is true. This Kripke structure is called a \emph{model} for $F$.
The above mentioned seriality axiom states that, if a formula holds in all reachable worlds, then there exists such a world. With the deontic reading of $\Box$ and $\Diamond$ this means whenever the formula $\Phi$ ought to be, then there exists a world where it holds. I.e. there is always a world in which the norms formulated by `the ought to be'-operator hold. 

To formalize the WST in deontic logic, we transform the statement about the cards into:
\begin{quote}
If there is a vowel on one side , it ought to be that the opposite side shows an even number.
\end{quote}
Using abbreviations $P$ and $Q$ this reads: $P \rightarrow \Box Q$\\
Now if we observe $P$, i.e. a card with a vowel on the upper side, we know that there is a world in which the deontic conditional holds, hence $Q$ holds. In other words  there is a world where the opposite side of the card contains an even number.

%\[P \rightarrow \Box Q\]
Note, that it would also be possible to formalize the statement as: $\Box ( P \rightarrow  Q)$.
In \cite{kutschera} there is a careful discussion which of these two possibilities should be used for conditional norms. The latter one has severe disadvantages, which is why we prefer the first method. %, where the $\Box$-operator is in the conclusion of the conditional. 
In Section~\ref{sec:consistency} we demonstrate that the alternative very easily results in an inconsistent normative system.

Assume for simplicity that the letters on the cards can only be $A, K$, the numbers only $4, 7$ and that we consider only one card. We represent the card by atoms of the form $c(l,A)$, $c(l,K)$, $c(n,4)$ and $c(n,7)$.
Where an $l$ in the first position denotes the letter side and an $n$ denotes the number side of a card. We further have formulae describing the way the card is constructed:
\begin{align}
	\top 					& \rightarrow c(l,A) \lor c(l,K). \label{equ:card1}\\
	c(l,A) \land c(l,K) 		& \rightarrow \bot. \label{equ:card2}\\
	\top					& \rightarrow c(n,4) \lor c(n,7).\label{equ:card3}\\
	c(n,4) \land c(n,7) 	& \rightarrow \bot. \label{equ:card4}
\end{align}
Formula~\eqref{equ:card1} states that the letter side of the card contains an $A$ or a $K$, formula~\eqref{equ:card2} states, that there is only one letter on the letter side of the card.  
Formulae~\eqref{equ:card3} and \eqref{equ:card4} describe the number side of the card respectively. 

The rule expressing the normative conditional reads in this simplified example as
\begin{equation}
 c(l, A) \rightarrow \Box c(n, 4) \label{equ:norm}
\end{equation}
Note that all the above formulae are propositional, although atoms like $c(l, A)$ seem to have a structure; logically, they are propositional variables being either $true$ or $\mathit{false}$. %In the sequel, we call a set of rules expressing the normative statements a \emph{normative system}.

%\vskip 2ex

\subsection{The  WST Task}
\label{wst-task}
Until now, we formalized the knowledge and the observation; we did not address a logical representation of the task itself. Then we want to use an automated reasoning system in order to solve the task and hence it is mandatory to query the system in a logical way.  To the best of our knowledge, we are not aware of such a formalization in the literature. Let's focus first on the abstract case without social or precaution context.

Usually, in logic based automated reasoning, a knowledge base $\mathit{KB}$ together with a query $Q$ is given and we want to know, if $Q$ is a logical consequence of $\mathit{KB}$, i.e. $\mathit{KB}\models Q$. In the WST the question is different, since it corresponds to a cheat detection task:
\begin{quote}
Given the knowledge $\mathit{KB}$, including the knowledge about norms, how can we detect cheating, or, which cards do we have to turn to detect a violation against the norm?
\end{quote}
In the sequel we use a standard tableau method for generating models. We assume the reader to be familiar with tableaux as introduced in \cite{modalLogic}.  %\remarkx{Reicht das als Einschr\"ankung auf WST aus?}
We don't use indexing of worlds because we treat the $\Box$-operator as a literal and do not expand it. In the examples of this paper this works, because we never have nested $\Box$-operators.
\begin{thesis}\label{thesis}Boxed literals occurring in open branches can be used for cheat detection: If for example an open branch contains literals $\Box F$ and $\Box G$ this branch tells us to check, if the current world fulfills both $F$ and $G$. 
\end{thesis}
Note that the information provided by an open branch are not necessarily minimal. %In our example this means, that there could be another open branch containing only the boxed literal $\Box F$ making it unnecessary to check if $G$ is fulfilled.
Therefore in order to find a minimal set of actions required for cheat detection, it is necessary to construct all open branches and to compare the set of boxed literals contained in the respective branches. Only those branches containing a minimal (w.r.t. set inclusion) set of boxed literals provide a minimal set of actions required for cheat detection.
%\begin{quote}
For the WST, this thesis leads to the following interpretation of open branches:
\begin{itemize}
\item If there is an open branch not containing any boxed literals, the observed situation does not require to check the hidden side of the card.
\item If all open branches contain the same boxed literal i.e. $\Box F$ we have to check the hidden side of the card (in the example, we have to check if  $F$ is fulfilled).
\item If all open branches contain boxed literals but not all open branches contain the same boxed literals, we have to compare the open branches with respect to the set of boxed literals. Those branches containing a minimal (w.r.t. set inclusion) set of boxed literals tell us what we have to check in order to make sure that the given norms are fulfilled.
\end{itemize}
%\end{quote}

\begin{thesis}
From a model-theoretic point of view turning a card to do cheat detection corresponds to the question, if there is a model for the set of formulae with a world fulfilling the observed situation which is a successor of itself. 
\end{thesis}
This ``self loop'' ensures, that whenever there is a boxed formula $\Box F$ which is true in the observed world $F$ has to be fulfilled as well. 
Intuitively this means, that this world corresponds to the observed situation and fulfills everything that ``ought to be''.
If there is no such model, it is obvious that the observed situation can only be caused by cheating.
 
% Next we present two ways to formalize this task and discuss the respective advantages and disadvantages.
 Next we discuss two formalizations of the WST.

\subsubsection{Naive Formalization}\label{naive}
The first formalization of the WST we present, consist of the set of formulae given in \eqref{equ:card1} to \eqref{equ:card4} together with the formula representing the norm given in formula~\eqref{equ:norm}. As an example, we add the observation of letter $A$ on the card. 
In the sequel, $\mathcal{B}$ denotes the set of formulae consisting of formula~\eqref{equ:card1} to \eqref{equ:card4} together with the observation and letter $\mathcal{N}$ denotes formula~\eqref{equ:norm}.

In Figure~\ref{fig:naive1} we give a tableau for the resulting set of formulae $\mathcal{B}\cup\mathcal{N}$ (as mentioned above, we do not expand the boxed formulae in the tableau). This tableau has two open branches:
%\remarkx{Die Beobachtung ist ja in B drin, stimmt so nicht ganz. Auch in den anderen Figures}
%\vspace{\baselineskip}

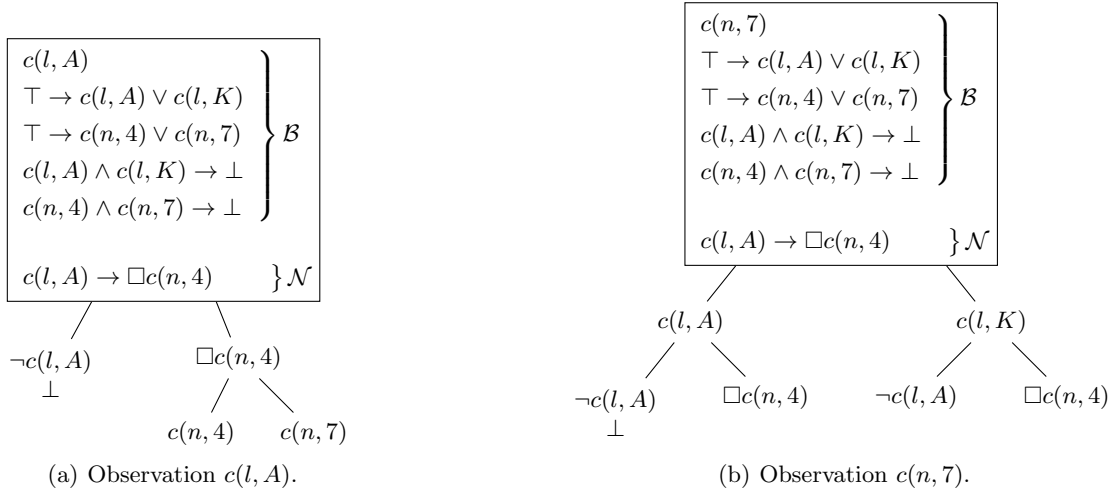
\begin{figure}[t]
\subfigure[ Observation $c(l,A)$. ]{\label{fig:naive1} 

\begin{tikzpicture}[level distance=3.5cm, sibling distance=3.0cm]
\node[name = n1,align=left, draw]{
 $\left.
  \begin{aligned}
	&c(l,A)\\
        &\top \rightarrow c(l,A) \lor c(l,K)\\
        &\top \rightarrow c(n,4) \lor c(n,7)\\
        &c(l,A) \land c(l,K) \rightarrow \bot\\
        &c(n,4) \land c(n,7)\rightarrow \bot \hspace{0.2cm}
     \end{aligned}
   \right\}\mathcal{B}
    $\\\\\\
 $\left.
  \begin{aligned}
	&c(l,A) \rightarrow \Box c(n,4)\hspace{0.4cm}
     \end{aligned}
   \right\}\mathcal{N}$
	}
    [sibling distance = 3.0cm, level distance=2.750cm] 
                   child{node {$\begin{array}{c} 
                   					 \neg c(l,A) \\
 							 \bot
						\end{array}$} [sibling distance = 2.0cm,level distance=1.0cm] 
                   			   }
                   child[sibling distance = 2.0cm,level distance=2.5cm]{node {$\Box c(n,4)$} child[sibling distance = 1.0cm,level distance=1cm]{node {$c(n,4)$}}
                   					   child[sibling distance = 2.0cm,level distance=1cm]{node{$c(n,7)$}}
                   };
%\node[right of=n2, node distance = 6cm, align = right]{$\mathcal{N}$  the normative world};
  \end{tikzpicture}

}\hfill
\subfigure[Observation $c(n,7)$.]{\label{fig:naive2}%\includegraphics[width=.45\linewidth]{wason-b1}
\begin{tikzpicture}
\node[name = n1,align=left, draw]{
    $\left.
  \begin{aligned}
	&c(n,7)\\
        &\top \rightarrow c(l,A) \lor c(l,K)\\
        &\top \rightarrow c(n,4) \lor c(n,7)\\
        &c(l,A) \land c(l,K) \rightarrow \bot\\
        &c(n,4) \land c(n,7)\rightarrow \bot \hspace{0.2cm}
     \end{aligned}
   \right\}\mathcal{B}
    $\\\\\\
 $\left.
  \begin{aligned}
	&c(l,A) \rightarrow \Box c(n,4)\hspace{0.4cm}
     \end{aligned}
   \right\}\mathcal{N}$        }
        [sibling distance = 15.0cm, level distance=2.5cm, node distance = 3cm] 
        child[sibling distance = 4cm, node distance = 3cm, level distance=2.5cm]{
            node {$c(l,A)$} 
        [sibling distance = 2cm, node distance = 3cm,level distance=1cm]
            child[level distance = 1.25cm]{
                node {$\begin{array}{c} \neg c(l,A) \\ \bot \end{array}$}
                }
            child{
                node{$\Box c(n,4)$}
                }
        }                   
        child[sibling distance = 4cm, node distance = 3cm, level distance=2.5cm]{
            node {$c(l,K)$}
        [sibling distance = 2cm, node distance = 3cm,level distance=1cm]
            child{
                node {$\lnot c(l,A)$}
                }
            child{
                node{$\Box c(n,4)$}
                }
        };
%\node[right of=n1, node distance = 6cm, align = right]{$\mathcal{B}$  the world of the observer\\ \\ \\\\\\\\the normative system $\mathcal{N}$};

% \node[right of=n2, node distance = 6cm, align = right]{$\mathcal{N}$  the normative world};
\end{tikzpicture}
}
\caption{Tableaux for the simplified 1-card WST with  naive formalization. $\mathcal{B}$  denotes the knowledge  of the observer and $\mathcal{N}$ the normative system.}
\label{fig:tab}
\end{figure}

%\qtreepadding 5mm
%%\qtreeunaryht 1cm
%\begin{figure}
%\begin{center}
%\begin{tikzpicture}[level distance=4.0cm, sibling distance=3.0cm]
%\node[name = n1,align=left, draw]{
%			  $ c(l,A)$\\
%                    	  $c(l,A) \lor c(l,K)\leftarrow \top$\\
%                     	  $c(n,4) \lor c(n,7)\leftarrow \top$\\
%                    	  $\bot \leftarrow c(l,A) \land c(l,K)$\\
%                    	  $\bot \leftarrow c(n,4) \land c(n,7)$\\
%                       $\Box c(n,4)\leftarrow c(l,A)$}
%    [sibling distance = 3.0cm, level distance=2.5cm] 
%                   child{node {$\begin{array}{c} 
%                   					 \neg c(l,A) \\
% 							 \bot
%						\end{array}$} [sibling distance = 2.0cm] 
%                   			   }
%                   child[sibling distance = 2.0cm]{node {$\Box c(n,4)$} child{node {$c(n,4)$}}
%                   					   child{node{$c(n,7)$}}
%                   };
%  \node[right of=n1, node distance = 6cm, align = right]{$\mathcal{B}$  the world of the observer\\ and the normative system $\mathcal{N}$};
%% \node[right of=n2, node distance = 6cm, align = right]{$\mathcal{N}$  the normative world};
%  \end{tikzpicture}
%  \caption{A tableau for the simplified 1-card Wason selection task for the naive formalisation and observation $c(l,A)$.}
%  \label{fig:naive1}
%\end{center}
%\end{figure}
\begin{align}
&B_{1} = \lbrace c(l,A), \Box c(n,4), c(n,4) \rbrace &
&B_{2} = \lbrace c(l,A), \Box c(n,4), c(n,7) \rbrace 
\end{align}\label{2-models}
Both open branches contain the same boxed literal $\Box c(n,4)$. According to Thesis~\ref{thesis}, this tells us to check, if the number side of the card depicts 4.
% in an open branch has the meaning, that it must be the case that in every reachable world $c(n,4)$ must be true. Because of the seriality of deontic logic, we know that there is a reachable world in which this norm, expressed by the boxed formula holds -- we just have to check whether we are in this ideal world. 

%The task for the test person is, to find the minimum number of cards which need to be turned in order to verify/falsify the statements given in the normative system. An occurrence of $\Box c(n,4)$ in an open branch means 'it ought to be the case, that the number side of the card shows the number 4'. 
%\remarkx{W\"are sch\"on, wenn man eine Intuition h\"atte, warum man nur umdrehen muss, wenn dei Box in allen %Modellen ist}
%If such a boxed literal occurs in a model, it can be therefore interpreted as 'we have to check, if the number side of the card shows number 4'. The boxed literals in each model tell us, which cards should be turned. If a certain boxed literal occurs in every model, we have to turn the respective card as a logical consequence. In our example, the boxed literal $\Box c(n,4)$ is contained in both models. Therefore we have to turn the card in oder to make sure that the statements in the normative system are fulfilled. 

Taking a closer look at the open branches reveals, that branch $B_{1}$ contains  $c(n,4)$ and $\Box c(n,4)$. In $B_{1}$, the number side of the card depicts 4 and it ought to be the case that the number side of the card depicts 4, meaning that $B_{1}$ fulfills the norm.
Contrary to that, $B_{2}$ contains $c(n,7)$ and $\Box c(n,4)$. In $B_{2}$, the number side of the card depicts 7 even though it ought to be the case that the number side of the card depicts 4. So $B_{2}$ violates the norm. Hence only from $B_{1}$ a model in form of  a Kripke structure containing a world fulfilling $\mathcal{B} \cup \mathcal{N}$ which has a ``self loop'' can be constructed.
%Note that by adding rules of the form $F \land \Box \lnot F \rightarrow \mathit{cheat}$ for $F \in \lbrace c(n,4), c(n,7), c(l,A), c(l,K)\rbrace$, we could represent cheat detection.

However this formalization of the WST does not always work as desired. Let us consider another example, where $7$ is observed on the number side of the card. The tableau for this example is given in Figure~\ref{fig:naive2}. This tableau has three open branches:
%
%\vspace{\baselineskip}
%\qtreepadding 5mm
%%\qtreeunaryht 1cm
%\begin{figure}
%\begin{center}
%\begin{tikzpicture}
%\node[name = n1,align=left, draw]{
%    $ c(n,7)$\\
%        $c(l,A) \lor c(l,K)\leftarrow \top$\\
%        $c(n,4) \lor c(n,7)\leftarrow \top$\\
%        $\bot \leftarrow c(l,A) \land c(l,K)$\\
%        $\bot \leftarrow c(n,4) \land c(n,7)$\\\\
%        $\Box c(n,4)\leftarrow c(l,A)$}
%        [sibling distance = 15.0cm, level distance=2.5cm, node distance = 3cm] 
%        child[sibling distance = 3.5cm, node distance = 3cm]{
%            node {$c(l,A)$} 
%        [sibling distance = 1.5cm, node distance = 3cm]
%            child{
%                node {$\begin{array}{c} \neg c(l,A) \\ \bot \end{array}$}
%                }
%            child{
%                node{$\Box c(n,4)$}
%                }
%        }                   
%        child[sibling distance = 3.5cm, node distance = 3cm]{
%            node {$c(l,K)$}
%        [sibling distance = 1.5cm, node distance = 3cm]
%            child{
%                node {$\lnot c(l,A)$}
%                }
%            child{
%                node{$\Box c(n,4)$}
%                }
%        };
%\node[right of=n1, node distance = 6cm, align = right]{$\mathcal{B}$  the world of the observer\\ \\ \\\\\\\\the normative system $\mathcal{N}$};
%
%% \node[right of=n2, node distance = 6cm, align = right]{$\mathcal{N}$  the normative world};
%\end{tikzpicture}
%  \caption{A tableau for the simplified 1-card Wason selection task for the naive formalisation and observation $c(n,7)$.}
%  \label{fig:naive2}
%\end{center}
%\end{figure}
\begin{align}
&B'_{1} = \lbrace c(n,7), c(l,A), \Box c(n,4) \rbrace &
&B'_{2} = \lbrace c(n,7), c(l,K), \lnot c(l,A) \rbrace \\
&B'_{3} = \lbrace c(n,7), c(l,K), \Box c(n,4) \rbrace &
\end{align}
In the case of observing 7 on the number side of the card, the desired conclusion is, that there has to be a $K$ on the letter side of the card. Hence we would expect to see $\Box c(l,K)$ in every open branch. However none of the open branches contains $\Box c(l,K)$. Taking a closer look at $\mathcal{B} \cup \mathcal{N}$ reveals, that it is not possible to deduce $\Box c(l,K)$ from this set. Making the even worse, is that it is not possible to deduce information on what ought to be depicted on the letter side of a card!

The reason for this is well known in the literature about deontic conditionals. With  a classical implication $ c(l,A) \rightarrow c(n,4)$ we can equivalently formulate the contrapositive $\lnot c(n,4) \rightarrow \lnot c(l,A)$ expressing, if there is not a 4 on the number side, there is no A on the letter side. In deontic logic, however, the norm is represented by $c(l,A) \rightarrow \Box c(n,4)$. The respective contrapositive is  $\lnot \Box c(n,4) \rightarrow \lnot c(l,A)$ or equivalently,  $\Diamond \lnot  c(n,4) \rightarrow \lnot c(l,A)$.
However what we want to state is: if we don't see a 4 on the number side, then there ought to be no A on the letter side.
This would be formalized as $\lnot  c(n,4) \rightarrow \Box \lnot c(l,A)$. Unfortunately this is not included in the naive formalization.  
Therefore we need to find a different formalization of the problem.

\subsubsection{Formalization using Pseudo-Contraposition}\label{contrapositive}
The drawback of the naive formalization of the WST is the fact, that it is not possible to deduce what ought to be the case for the letter side of the card. To remedy this situation, we use a second norm called \emph{pseudo-contrapositive}:
\begin{equation}\label{equ:contrapositiv1}
\lnot  c(n,4) 	 \rightarrow \Box \lnot c(l,A)\end{equation}
which can be transformed into: 
%\begin{equation}\label{equ:contrapositiv2}
$c(n,7) \rightarrow \Box c(l,K)$
%\end{equation}

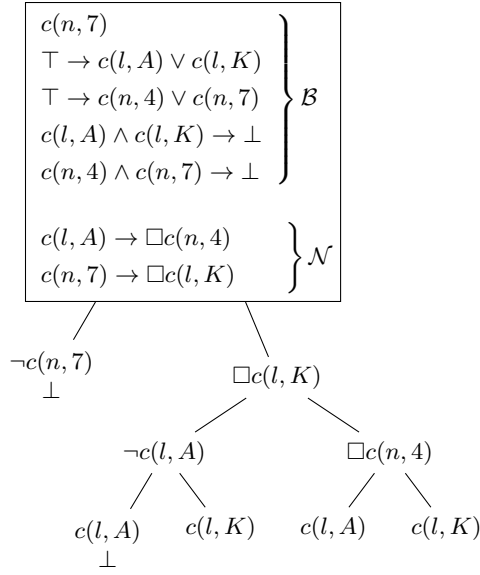
\begin{figure}[t]
\begin{center}
\begin{tikzpicture}
\node[name = n1,align=left, draw]{
    $\left.
  \begin{aligned}
       &c(n,7)\\
       &\top \rightarrow c(l,A) \lor c(l,K)\\
       &\top  \rightarrow c(n,4) \lor c(n,7)\\
       &c(l,A) \land c(l,K) \rightarrow  \bot\\
       &c(n,4) \land c(n,7)  \rightarrow \bot\hspace{0,2cm}  
       	\end{aligned}
   \right\}\mathcal{B}
    $\\\\\\
 $\left.
  \begin{aligned}
       &c(l,A) \rightarrow \Box c(n,4)\\
       &c(n,7) \rightarrow \Box c(l,K)\hspace{0,66cm}  
  	\end{aligned}
   \right\}\mathcal{N}
    $

}
        [sibling distance = 15.0cm, level distance=3.0cm, node distance = 3cm] 
        child[sibling distance = 3.5cm, node distance = 3cm]{
            node {$\begin{array}{c} \neg c(n,7) \\ \bot \end{array}$} 
        }                   
        child[sibling distance = 2.5cm, level distance=3.0cm, node distance = 3cm]{
            node {$\Box c(l,K)$}
        [sibling distance = 3.0cm, node distance = 3cm, level distance=1cm]
            child{
                node {$\lnot c(l,A)$}
                        [sibling distance = 1.5cm, node distance = 3cm, level distance=1cm]
       		     child[level distance = 1.25cm]{
	                	node {$\begin{array}{c} c(l,A) \\ \bot \end{array}$}
                    }
            		child{
                		node{$c(l,K)$}
                	}
                }
            child{
                node{$\Box c(n,4)$}
                [sibling distance = 1.5cm, node distance = 3cm,level distance=1cm]
       		     child{
	                	node {$c(l,A)$}
                    }
            		child{
                		node{$c(l,K)$}
                	}
                }
        };
%  \node[right of=n1, node distance = 4cm, align = right]{$\mathcal{B}$  \\\\\\\\\\\ $\mathcal{N}$};

% \node[right of=n2, node distance = 6cm, align = right]{$\mathcal{N}$  the normative world};
\end{tikzpicture}
\end{center}
\caption{Tableaux for the simplified 1-card WST with for the formalization using the pseudo-contrapositive. $\mathcal{B}$  denotes the knowledge  of the observer and $\mathcal{N}$ the normative system.}\label{fig:contrapositiv}
\end{figure}

We add this norm to the naive formalization resulting in the new normative system:
\begin{align}
c(l,A) 		& \rightarrow \Box c(n,4)\label{equ:norm1}\\
c(n,7) 		& \rightarrow \Box c(l,K)
\end{align}
%We call the formula  in Equation~\eqref{equ:contrapositiv2} \emph{pseudo-contrapositive}.
With the of help the pseudo-contrapositive, we are now able to calculate a solution for the previous example. Again, we observe card $c(n,7)$. Fig.~\ref{fig:contrapositiv} shows the tableau for the resulting set of formulae. This tableau has three open branches:%. We give the models for those branches:
\begin{align}
&B''_{1} = \lbrace c(n,7), \Box c(l,K), \lnot c(l,A), c(l,K) \rbrace &
&B''_{2} = \lbrace c(n,7), \Box c(l,K), \Box c(n,4), c(l,A) \rbrace \\
&B''_{3} = \lbrace c(n,7), \Box c(l,K), \Box c(n,4), c(l,K) \rbrace &
\end{align}

All three branches contain $\Box c(l,K)$, so we can deduce, that the letter side of the card ought to show a $K$.
\subsubsection{Reducing the Wason Selection Task to a Satisfiability Test}%\remarkx{Eigentlich m\"ussen wir immer erst einen Erf\"ullbarkeitstest machen. W\"are z.B. c(l,A) und c(n,7) gegeben, w\"are das schon unerf\"ullbar und dann k\"onnte man hinzuf\"ugen was man will und h\"atte immer eine unerf\"ullbare Formelmenge.}
As mentioned before, the question in the WST is to detect cheating or to find out, which cards have to be turned in order to detect a violation of the norm. 
The formalization using Pseudo-Contraposition presented in \ref{contrapositive} can be used to detect, if a card has to be turned.

Next we will transform this question into a satisfiability test: If a card has to be turned, this information is contained in all models. Given e.g. the observation $c(n,7)$, all models constructed contained $\Box c(l,K)$, meaning that the letter side of the card ought to show $K$. Assuming that the set of formulae under consideration is satisfiable, another possibility would be to add $\lnot \Box c(l,K)= \Diamond c(l,A)$ to the set of formulae and show that the resulting set of formulae is unsatisfiable.

In the next section, we will use an automated theorem prover to solve this satisfiability test. This leads us to an automated solution of the question of the WST.

\subsection{WST and Automated Theorem Proving}\label{hyper}
SDL can be translated into decidable fragments of first order logic. See \cite{SH13} for details. Hence practically any first order theorem prover could be used to reason in SDL.

Hyper \cite{WernhardPelzer} is a theorem prover for first order logic with equality. It is the implementation of the  E-hypertableau calculus \cite{baumgartnerFurbachPelzer} which extends the hypertableau calculus with superposition based equality handling. Hyper has been successfully used in various AI-related applications, like intelligent interactive books or natural language query answering.  
One of the advantages of the hyper tableau calculus is the avoidance of unnecessary or-branching. This is one reason why we decided to use Hyper to reason in SDL.
Another reason is the fact, that recently the  E-hypertableau calculus and its implementation have been extended to deal with knowledge bases given in the description logic $\mathcal{SHIQ}$ \cite{cadesd}. There is a strong connection between modal logic and description logic. As shown in \cite{Schild91acorrespondence},  the description logic $\mathcal{ALC}$ is  a notational variant of the modal logic $\mathcal{K}_{n}$. Therefore any formula given in the modal logic $\mathcal{K}_{n}$ can be translated into an $\mathcal{ALC}$ concept and vice versa. When using Hyper as a theorem prover for SDL, it is not necessary to translate the SDL formulae into first order logic. It is sufficient to translate them to description logic which is more closely related to SLD then first order logic.%\remarkx{Ziemliches Blabla. Verbesserungsw\"urdig.}
Since we are only considering a modal logic as opposed to a multimodal logic, we will omit the part of the translation handling the multimodal part of the logic.
Table~\ref{tab:translationmodal1} gives the inductive definition of a mapping $\phi$ from modal logic $\mathcal{K}$ formulae to $\mathcal{ALC}$ concepts. \begin{table}[t]
\centering
\begin{tabular}{rclcrcl}
$\phi(\top)$ 			& = &  $\top$&\hspace{5ex}& $\phi(\bot)$ 			& = &  $\bot$\\
$\phi(a)$ 				& = &  $a$&&$\phi(\lnot c)$ 			& = & $\lnot \phi(c)$\\
$\phi(c \land d)$ 		& = & $\phi(c) \sqcap \phi(d)$&\hspace*{10ex}&$\phi(c \lor d)$ 		& = & $\phi(c) \sqcup \phi(d)$\\
$\phi(\Box c)$ 			& = & $\forall r.\phi(c)$&&$\phi(\Diamond c)$ 	& = & $\exists r.\phi(c)$
\end{tabular}

\caption{Translation of modal logic $\mathcal{K}$ formulae into $\mathcal{ALC}$ concepts.}
\label{tab:translationmodal1}
\end{table}
%\remarkx{Brauchen wir hier Syntax und Semantik von $\mathcal{ALC}$?}

Mapping $\phi$ can be used to translate the deontic logic formulae into the description logic $\mathcal{ALC}$ as well. %Later Hyper can be used to check if the observed world obeys the given normative system.  
%In this paper we address two different tasks. In the first reasoning task, we are given the world of an observer together with a normative system. 
%\remarkx{Ist das consistent mit der Behandlung vorher; wo M1 und M2 konstruiert wurden?}
%After translating the set of deontic logic formulae used in the framed part of Fig.~\ref{fig:contrapositiv} into $\mathcal{ALC}$\footnote{Further in line \eqref{tab:line6} to \eqref{tab:line9} of Table~\ref{tab:translation} we have to add formulae describing the way the cards are constructed for all reachable worlds.} a satisfiability test is used  tofind out if a card has to be turned. %This is why we have to the formulae given in \eqref{f5} - \eqref{f8}. 
The result of the translation of all formulae is shown in Table~\ref{tab:translation}.
For readability reasons we decided to keep the arguments of a ground atom e.g. $c(l,A)$ in the concept introduced for the atom. Therefore we translate atoms like $c(l,A)$ into  atomic concepts $c(l,A)$. 
\begin{table}[t]
\begin{center}
\begin{tabular}{l | lc}
\textbf{Deontic Logic} & $\mathcal{ALC}$ &\\
\hline
%	$ c(n,7)$ 							& \phantom{$\lnot$}$c(n,7)$  				&{\refstepcounter{rowcount}\label{tab:line1}\hspace*{\tabcolsep}(\therowcount)\hspace*{\tabcolsep}}	\\\\
$\begin{aligned}	
	\top & \rightarrow  c(n,7)\\
	\top  & \rightarrow c(l,A) \lor c(l,K)\\
	c(l,A) \land c(l,K) &\rightarrow \bot \\
	\top	&\rightarrow c(n,4) \lor c(n,7)\\
	c(n,4) \land c(n,7) &\rightarrow \bot\\
	\top  &\rightarrow \Box(c(l,A) \lor c(l,K))\\
	\Box(c(l,A) \land c(l,K)) & \rightarrow \bot\\
	\top	&\rightarrow \Box (c(n,4) \lor c(n,7))\hspace{2em}\\
	\Box(c(n,4) \land c(n,7)) &\rightarrow \bot\\
	c(l,A) &\rightarrow \Box(c(n,4))\\
	c(n,7) &\rightarrow \Box(c(l,K))\\
	\Box \Phi &\rightarrow \Diamond \Phi
\end{aligned}$
				 		& 
$\begin{aligned}	
	& \phantom{\lnot}c(n,7)\\
	\hspace{2em}&\phantom{\lnot}c(l,A) \sqcup c(l,K)		\\
	& \lnot c(l,A) \sqcup \lnot c(l,K)			\\
	&\phantom{\lnot}c(n,4) \sqcup c(n,7)		\\
	& \lnot c(n,4) \sqcup \lnot c(n,7)		\\
	& \phantom{\lnot}\forall r.(c(l,A) \sqcup c(l,K))\\
	& \forall r.(\lnot c(l,A) \sqcup \lnot c(l,K))\\
	& \phantom{\lnot}\forall r.(c(n,4) \sqcup c(n,7))\\
	& \forall r.(\lnot c(n,4) \sqcup \lnot c(n,7))\\
	&\lnot c(l,A) \sqcup  \forall r.(c(n,A))\\
	&\lnot c(n,7) \sqcup \forall r.(c(l,K))\\
	&\top \sqsubseteq \exists r.\top
	\end{aligned}$
	&
	$\begin{aligned}
	&\phantom{\forall}{\refstepcounter{rowcount}\label{tab:line1}\hspace*{\tabcolsep}(\therowcount)\hspace*{\tabcolsep}}\\
	&\phantom{\forall}{\refstepcounter{rowcount}\label{tab:line2}\hspace*{\tabcolsep}(\therowcount)\hspace*{\tabcolsep}}\\
	&\phantom{\forall}{\refstepcounter{rowcount}\label{tab:line3}\hspace*{\tabcolsep}(\therowcount)\hspace*{\tabcolsep}}\\
	&\phantom{\forall}{\refstepcounter{rowcount}\label{tab:line4}\hspace*{\tabcolsep}(\therowcount)\hspace*{\tabcolsep}}\\
	&\phantom{\forall}{\refstepcounter{rowcount}\label{tab:line5}\hspace*{\tabcolsep}(\therowcount)\hspace*{\tabcolsep}}\\
	&\phantom{\forall}{\refstepcounter{rowcount}\label{tab:line6}\hspace*{\tabcolsep}(\therowcount)\hspace*{\tabcolsep}}\\
	&\phantom{\forall}{\refstepcounter{rowcount}\label{tab:line7}\hspace*{\tabcolsep}(\therowcount)\hspace*{\tabcolsep}}\\
	&\phantom{\forall}{\refstepcounter{rowcount}\label{tab:line8}\hspace*{\tabcolsep}(\therowcount)\hspace*{\tabcolsep}}\\
	&\phantom{\forall}{\refstepcounter{rowcount}\label{tab:line9}\hspace*{\tabcolsep}(\therowcount)\hspace*{\tabcolsep}}\\
	&\phantom{\forall}{\refstepcounter{rowcount}\label{tab:line10}\hspace*{\tabcolsep}(\therowcount)\hspace*{\tabcolsep}}	\\
	&\phantom{\forall}{\refstepcounter{rowcount}\label{tab:line11}\hspace*{\tabcolsep}(\therowcount)\hspace*{\tabcolsep}} \\
	&\phantom{\forall}{\refstepcounter{rowcount}\label{tab:line12}\hspace*{\tabcolsep}(\therowcount)\hspace*{\tabcolsep}}		
	\end{aligned}$
\end{tabular}
\end{center}
\caption{Translation of formulae given in the framed part of Figure~\ref{fig:contrapositiv} into $\mathcal{ALC}$.}
\label{tab:translation}
\end{table}

Note that line \eqref{tab:line1} of Table~\ref{tab:translation} describes the world we observe. Further line \eqref{tab:line2} to \eqref{tab:line5} describe the way the cards are constructed. The construction of the cards should be effective for all reachable worlds. This is why we add the formulae given in line \eqref{tab:line6} to \eqref{tab:line9}. The conjunction of those formulae are denoted by $\mathcal{B}$. $\phi(\mathcal{B})$ denotes the result of the translation into an $\mathcal{ALC}$ concept.
%So 
%\begin{equation}
%\phi(\mathcal{B}) = c(l,A) \sqcap (c(l,A) \sqcup c(l,K) \sqcap (\lnot c(l,A) \sqcup \lnot c(l,K)) \sqcap (c(n,4) \sqcup c(n,7)) \sqcap (\lnot c(n,4) \sqcup \lnot c(n,7))
%\end{equation}
Line \eqref{tab:line10} and \eqref{tab:line11} describe the norm $\mathcal{N}$ and line \eqref{tab:line11} presents the translation of the seriality axiom.
%\begin{equation}
%\phi(\mathcal{N}) = (\lnot c(l,A) \sqcup  \forall r.(c(n,4))) \sqcap  (\lnot c(n,7) \sqcup  \forall r.(c(l,K)))
%\end{equation}
%
%In addition to that we have to translate the seriality axiom into description logic. In \cite{Klarman28052013} it is shown, that the seriality axiom can be translated into the following TBox:
%\begin{equation*}
%\mathcal{T}= \lbrace \top \sqsubseteq \exists r.\top\rbrace
%\end{equation*}
%with $r$ the atomic role introduced by the above described translation. The translation of the seriality axiom is shown in the last line of Table~\ref{tab:translation}.

Now the  theorem prover Hyper is used to calculate if the card has to be turned in order to find out, if the observed situation obeys the given normative system. For this, the $\mathcal{ALC}$ concepts given in the right column of Table~\ref{tab:translation} are translated into DL-clauses, which is the input language of Hyper. We denote this transformation by $\Xi$. The set of DL-clauses for the right column of Table~\ref{tab:translation} is
\begin{equation}
\Xi(\phi(\mathcal{B}) \cup \phi(\mathcal{N}) \cup \mathcal{T})
\end{equation}
During the transformation into DL-clauses, many auxiliary concepts are introduced which makes the resulting set of DL-clauses complicated to read. Since the DL-clauses are not important to understand our example, we refrain from presenting them. See \cite{msh07optimizing} for details on DL-clauses.

%The previous section described how to transform the WST into a satisfiability test. Now this technique is used to find out if the card has to be turned. 
In order to check, if we have to turn the card, the DL-clauses for the concept $\lnot \forall r.c(l,K)$ are added to the set of DL-clauses and afterwards Hyper is used to check the satisfiability of the resulting set. According to Hyper, the resulting set of DL-clauses
\begin{equation*}
\Xi(\phi(\mathcal{B}) \cup \phi(\mathcal{N}) \cup \mathcal{T} \cup \lbrace \lnot \forall r.c(l,K)\rbrace)
\end{equation*}
is unsatisfiable. Hence we know that we have to turn the  card. 

Since there is no TBox in deontic logic, the translation of the formulae given in Table~\ref{tab:translation} lead to a description logic concept together with one TBox axiom for the seriality axiom. The seriality axiom has to be added to the TBox, because it is supposed to be true for every  word. 
Another possibility to formalize the WST  would be to directly use description logic and to use the TBox not only for the seriality axiom. The formulae describing the way the cards are constructed are also supposed to be true in every reachable world. Hence it makes sense to add the translation of those formulae into the TBox.  This leads to the following TBox:
\begin{align} 
\mathcal{T} = \lbrace	& \top \sqsubseteq \exists r.\top,\\
					& \top \sqsubseteq c(l,A) \sqcup c(l,K),\\
					& \top \sqsubseteq \lnot c(l,A) \sqcup \lnot c(l,K),\\
					& \top \sqsubseteq c(n,4) \sqcup c(n,7),\\
					& \top \sqsubseteq \lnot c(n,4) \sqcup \lnot c(n,7) \rbrace
\end{align} 
Note that, since the TBox is true in all  worlds, we do not have to add formulae corresponding to line \eqref{tab:line6} to \eqref{tab:line9} of Table~\ref{tab:translation} to the TBox.
%For the solution of WST, we have to check, if the observed world obeys the given normative system. This corresponds to checking the satisfiability of $\mathcal{B} \land \mathcal{N}$ or in $\mathcal{ALC}$ the satisfiability of $\phi(\mathcal{B}) \sqcap \phi(\mathcal{N})$.

%\remarkx
%\remarkx{Vielleicht die 2. Task in einen spaeteren Abschnitt verschieben?}

\subsection{Abstract vs context WST}\label{sect:abstract vs context WST}

We modeled the WST in deontic logic and discussed the use of an automated theorem prover to compute cheat detection.
Next we address how the differences of the performance of humans in the abstract and in the context case of  the WST can be modeled with the help of our approach.

We argued in Section~\ref{wst-task} that the formalization of the task resulted in a check whether the open branches all contain the same  boxed  literal, as it was the case in the  branches  $B_1$ and $B_2$ on page~\pageref{2-models}. Such an occurrence of an 'ought-to'-literal tells us that it has to be checked.

%\remarkx{Hier jetzt ein Bezugg auf die Proof task von vorhin}

%
% in Equation~\eqref{card1} for card 1. From this we  generated the  proof task, as given in equation \eqref{wst-card1}, \eqref{wst-card1-taut} or \eqref{wst-card1-unsat}. 
 
 Solving the WST in this abstract case more or less makes it necessary to involve a logical calculus  as done by Hyper  in order to construct the models and to check it with respect to the boxed literals -- obviously humans are not good at constructing models out of the given specification.

In the case of a context, we follow the hypothesis that humans have the appropriate models explicitly in their mind. They have been constructed by prior experience or even by evolution. This is very much in accordance with the mental model theory from Johnson-Laird, which is elaborated for the case of conditional reasoning in \cite{johnson2002conditionals}. There the authors assume, that there is a mental  representation of models for conditionals as they are used in the WST. It is argued, that the form and nature of the representation heavily influences the performance of people solving WST. 
\begin{thesis}
In the case of a social contract or a precaution rule, humans have the models of a world in which the norms hold in form of an explicit  mental representation ready at hand.  There is no need to construct them like it was necessary in the abstract case. --  They just have to compare the observations in the WST with their mental model.
\end{thesis}
 %Please note, that this explicit model construction is not necessary for the entire logical specification, only those clauses which use the deontic operator $\Box$ to denote a norm can be handled by an explicit model.
%\medskip
To sum up, in both  the abstract and the context version of the WST,  we have  a \emph{model checking task}. In the abstract case the model is given only implicitly by the rules for the norm --- before comparing it, it has to be constructed. This can be done by a logical calculus, just as we demonstrated with the Hyper-prover. If all models from the result of the prover still contain the same boxed literal, we have to check it. This construction obviously is error prone if carried out by  humans.

\medskip

\begin{figure}[t]
\begin{center}
\begin{tikzpicture}[level distance=1cm]
\node[name = n1,align=left, draw]{
    		$\begin{aligned}
			\mathit{under\_21} 	&\rightarrow \Box\lnot (\mathit{drink\_beer})\\
        			\mathit{drink\_beer} 	&\rightarrow \Box\lnot (\mathit{under\_21})
		\end{aligned}$
		}
        [sibling distance = 15.0cm, node distance = 3cm] 
        child[sibling distance = 6.5cm, node distance = 3cm]{
            node {$\lnot \mathit{under\_21}$} 
        [sibling distance = 2.5cm, node distance = 3cm]
            child{
                  node{$\lnot \mathit{drink\_beer}$}                }
            child{
                node{$\Box\lnot (\mathit{under\_21})$}
                }
        }                   
        child[sibling distance = 3.5cm, node distance = 3cm]{
            node {$\Box\lnot (\mathit{drink\_beer})$}
        [sibling distance = 2.5cm, node distance = 3cm]
            child{
 		node{$\lnot \mathit{drink\_beer}$}
                }
            child{
                node{$\Box\lnot (\mathit{under\_21})$}
                }
        };
%\node[right of=n1, node distance = 6cm, align = right]{$\mathcal{B}$  the world of the observer\\ \\ \\\\\\\\the normative system $\mathcal{N}$};

% \node[right of=n2, node distance = 6cm, align = right]{$\mathcal{N}$  the normative world};
\end{tikzpicture}
\caption{Tableaux for the two norms for the social contract version of the WST.}
\label{fig:socialcontract}
\end{center}
\end{figure}
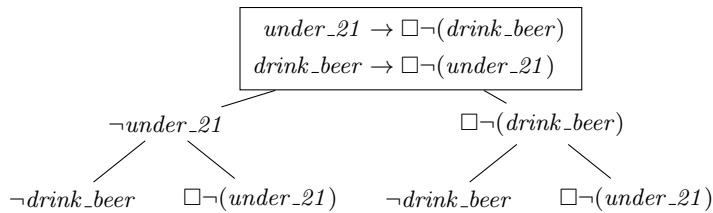

In the remainder of this section we will work this out in detail, with the help of the experiment given in  part (b) of Figure~\ref{fig:wason}.  The social contract rule for this example could be
\begin{equation} 
\mathit{under\_21} \rightarrow \Box\lnot \mathit{drink\_beer}
\end{equation}
We add the pseudo-contrapositive  to this  formalization
\begin{equation} 
\mathit{drink\_beer} \rightarrow \Box\lnot \mathit{under\_21}
\end{equation}
resulting in the two formulae of the framed part in Figure~\ref{fig:socialcontract}.
This tableau has four  open branches:
\begin{align}
B_{1} & = \lbrace \lnot \mathit{under\_21}, \lnot \mathit{drink\_beer}\rbrace &
B_{2} & = \lbrace \lnot \mathit{under\_21}, \Box \lnot \mathit{under\_21}\rbrace \label{sc2}\\
B_{3} & = \lbrace \Box \lnot \mathit{drink\_beer}, \lnot \mathit{drink\_beer}\rbrace &
B_{4} & = \lbrace \Box \lnot \mathit{drink\_beer}, \Box \lnot \mathit{under\_21}\rbrace \label{sc4}
\end{align}
Branches $B_{1}$ and $B_{2}$ represent those cases, in which the observed persons age is over 21. Both models do not contain a boxed literal concerning the beverage. Therefore, whenever we observe a person of age clearly over 21, we instantly know, that we do not have to take a closer look at the beverage.
This is totally different, as soon as the observed person is younger than 21. This case contradicts $B_{1}$ and $B_{2}$. This is why we have to consider $B_{3}$ and $B_{4}$ in this case. Both $B_{3}$ and $B_{4}$ contain $\Box \lnot \mathit{drink\_beer}$, stating that the observed person is not allowed to drink beer. Therefore we  know, that we have to check the beverage. 
%\remarkx{Was genau soll in die Thesis Umgebung?}

We argued above, that in the social contract case, the models are already at hand and  just have to be compared with the open branches.  Those four branches are already constructed as mental models in our brain. When we are confronted with the social contract version of the selection task, we don't have to perform the error prone construction of those models. We can use the mental models we have at hand and therefore we are able to perform the social contract version of the WST much better then the abstract version.

%Assume we observe a card which states $\mathit{under\_21}$, then we know imediately that $M_1$ and $M_2$ both are no models. For  $M_3$ and $M_4$ we have to assure that  $\Box \lnot (\mathit{drink\_beer})$, which is contained in both, is true. 

%\remarkx{Hier fehlt mir weider die Intuition - wie vorhin schon mal angemahnt :-(}
In the other case, where we observe a person drinking beer the two branches which remain are $B_2$ and $B_4$, where we have to make sure that $\Box \lnot \mathit{under\_21}$ holds. Or put it differently, we have learned that the only cases where we have to test are those where the premises of our norm and its pseudo-contrapositive holds.

%
%In our example, the models $M =\lbrace \lnot \mathit{under\_21} \rbrace$ and $M' = \lbrace \Box \lnot (\mathit{drink\_beer}) \rbrace$ are known from experience. Model $M$ can be interpreted as ``The social contract rule is not violated, as long as the observed person is not under 21.''.  The other model tells us to make sure, that the observed person does not drink beer. Confronted with the cards given in Figure~\eqref{fig:wason}, the test person compares it with $M$ and realizes, that age 22 is consistent with model $M$. Hence the test person knows instantly, that she does not have to turn the card containing the number 22. Since the card displaying number 16 is not compatible with model $M$, it is clear that we have to turn this card in order to make sure that the social contract rule is not violated.
%\medskip
%
%\remarkx{Hier müsste wohl noch ein Beispiel dafür hin, dass das Model explizit gegeben  ist?}

Our approach using deontic logic can be easily extended to handle the effect of the patient from \cite{cosmides2008can}. This person had a severe brain damage such that he  was able to solve the precaution task very well, but  in the task with the social contract he performed as badly as in the abstract case. It seems as if the mental representation of a model for the norms concerning precaution rules still exist, while the model of the social contract norm disappeared, it has to be constructed very much like in the abstract case.

To model such a behavior we only have to switch to multi-modal logic; instead of one ought-to operator $\Box$, we simple introduce an operator $\Box_{sc}$ for social contract norms and another $\Box_{pr}$ for precaution rules. Hence we could formulate conditionals with different contexts with different modal operators:
The social contract rule from our example in Figure~\ref{fig:wason} could be 
\[ \mathit{under\_21} \rightarrow \Box_{sc}\lnot (\mathit{drink\_beer})
\]
whereas  a  precaution rule could be
\[\mathit{driving\_a\_car}  \rightarrow  \Box_{pr} (\mathit{fasten\_seatbelt})
\]

With such a multi-modal logic it  could be that a reasoner has an explicit representation of a norm expressed with one operator, while for the other operator it has to compute the model, before solving the task, just the same way as in the abstract case.

We suggested above, to use the Hyper theorem prover for reasoning in deontic logic. Hyper is able to handle knowledge bases given in the description logic $\mathcal{SHIQ}$, which is the description logic $\mathcal{ALC}$ extended with transitive roles, role hierarchies, qualified number restrictions and inverse roles. 
%\remarkx{Bisher war immer nur von ALC die Rede?}
Since $\mathcal{SHIQ}$ allows the usage of more than one role, Hyper can be used to reason in multi-deontic logic as well.
%\remarkx{Bemerkung zur multi-deontic in Hyper}

\section{Deontic Logic and the Suppression task}
\label{sec:suppression}
\newcommand{\essay}{\ensuremath \mathit{essay\_to\_write}}
\newcommand{\study}{\ensuremath study\_late}
\newcommand{\textbook}{\ensuremath textbooks\_to\_read}
\newcommand{\library}{\ensuremath library\_open}

Another well researched phenomena is the  \emph{suppression task}. In \cite{Byrne198961} a series of experiments are reported, which demonstrate that human reasoning is non-monotonic in a certain sense. Given the following two statements:
\begin{quote}
If she has an essay to write she will study late in the library.\\
She has an essay to write.
\end{quote}
In an experiment persons are asked to draw a valid conclusion out of these premisses, it turned out that 98\%
of the test persons conclude correctly that
\begin{quote}
She will study late in the library.
\end{quote}
This shows that in such a setting modus ponens a is very natural rule of deduction.
If an additional statement is given, namely 
\begin{quote}
If she has some textbooks to read she will study late in the library.
\end{quote}
this does not change the percentage of correct answers. Obviously this additional conditional is understood as an alternative.
And indeed, we can transform the two conditionals
\begin{eqnarray*}
\essay \rightarrow  \study\\
\textbook \rightarrow \study
\end{eqnarray*}
equivalently into a single one, where the premise is an disjunction:

\[\essay \lor \textbook \rightarrow \study \]

If however as an  additional premiss 
\begin{quotation}
If the library stays open she will study late in the library.
\end{quotation}
or as a formula $\library \rightarrow \study $
is added, only 38\% draw the correct conclusion, although modus ponens is applicable in this case as well. People are understanding this additional conditional not as an alternative but as an additional premiss. 

We propose the same method as applied in the case of the WST. The conditional $\library \rightarrow \study$ is not just additional knowledge, moreover it can be understood as  trigger of additional knowledge about the world. We know that  usually to study  late in the library, the  library is  open  $\study\rightarrow  \library$.

If we assume his additional formula as a  \emph{norm}; which can be  formulated  with the help of the deontic ought-to-operator $\Box$, this leads to the following:

\begin{align}
\essay 	\rightarrow  &\study\label{cond-essay}\\
\library 	\rightarrow &\study \\
\study 	\rightarrow &\Box    \library\label{norm}\\
		&\essay \label{essay}
\end{align}

The question is $\study$, can easily be answered positively by using formulae \eqref{cond-essay} and \eqref{essay}. If, however, the norm \eqref{norm} is taken into account, the questions corresponds to a model checking task as discussed in Section~\ref{wst-task}. We can easily find a model 
\[M = \{\essay, \study, \Box \library\}\]
  by constructing a tableau similar to the one in Figure~\ref{fig:contrapositiv}. However we are not able to check -- in contrast to the WST -- whether $\Box \library$ holds, it \emph{ought to be the case}. This explains why much lesser persons are answering the question, whether she is studying late positively.

\section{Consistency Testing of Normative Systems}
\label{sec:consistency}
% \remarkx{Definition: Normative System fehlt!}

In the philosophical literature deontic logic is also used to formulate entire normative systems (e.g. \cite{kutschera}).
In practice such normative systems can be rather complex. This makes it difficult for the creator of a normative system to see if a normative system is consistent. We will show that it is helpful to be able to check consistency of normative systems automatically. We will use the Hyper theorem prover for this task.

As an example, we consider the well-known problem of \emph{contrary-to-duty obligations} introduced in \cite{chisolm}. In Table~\ref{tab:contrary-to-duty} the problem is given in natural language together with its formalization in deontic logic.
\begin{table}[t]
\begin{center}
\begin{tabular}{l | c}
\textbf{Natural Language} & Normative System $\mathcal{N}'$ \\
\hline
 $a$ ought not steal.										&  $\Box \lnot s$\\
 $a$ steals.												& $s$\\
 If $a$ steals, he ought to be punished for stealing.				&$s \rightarrow \Box p$\\
 If $a$ does not steal, he ought not be punished for stealing.	& $\Box (\lnot s \rightarrow \lnot p)$
\end{tabular}
\end{center}
\caption{Contrary-to-duty obligation together with the formalization in deontic logic.}
\label{tab:contrary-to-duty}
\end{table}
As shown in \cite{kutschera}, the normative system given in Table~\ref{tab:contrary-to-duty} is inconsistent. 
We will use Hyper to show this inconsistency. For this, we first transform $\mathcal{N'}$ into $\mathcal{ALC}$. The result of this transformation is given in Table~\ref{tab:translation2}.
\begin{table}[t]
\begin{center}
\begin{tabular}{l |l  }
\textbf{Deontic Logic} & $\mathcal{ALC}$ \\
\hline
$\Box \Phi \rightarrow \Diamond \Phi$& $\top \sqsubseteq \exists R.\top$	\\
							&						\\
$\Box \lnot s$ 					& $\forall R. \lnot S$ 			\\
$s$ 							&  $S$ 					\\
$s \rightarrow \Box p$ 			& $\lnot S \sqcup \forall R.P$	\\
 $\Box (\lnot s \rightarrow \lnot p)$	& $\forall R.(S \sqcup \lnot P)$	\\
							 \end{tabular}
\end{center}
\caption{Translation of the normative system $\mathcal{N}'$ into $\mathcal{ALC}$.}
\label{tab:translation2}
\end{table}
%\begin{table}[ht]
%\begin{center}
%\begin{tabular}{l |l | l}
%\textbf{Deontic Logic} & $\mathcal{ALC}$ & DL-Clauses\\
%\hline
%$\Box \Phi \rightarrow \Diamond \Phi$& $\top \sqsubseteq \exists R.\top$	& $\exists R.\top (X).$\\
%							&						&$b_{1}(a).$\\
%							&						& $b_{2}(X) \leftarrow b_{1}(X).$\\
%$\Box \lnot s$ 					& $\forall R. \lnot S$ 			& $\bot \leftarrow b_{2}(X)\land r(X,Y)\land s(Y)$\\
%$s$ 							&  $S$ 					& $s(X) \leftarrow b_{2}(X).$\\
%$s \rightarrow \Box p$ 			& $\lnot S \sqcup \forall R.P$	& $p(Y) \leftarrow b_{2}(X) \land s(X) \land r(X,Y).$\\
% $\Box (\lnot s \rightarrow \lnot p)$	& $\forall R.(S \sqcup \lnot P)$	& $b_{3}(Y) \leftarrow b_{2}(X) \land r(X,Y).$\\
%							&						& $s(X) \leftarrow b_{3}(X) \land p(X).$
% \end{tabular}
%\end{center}
%\caption{Translation of the normative system $\mathcal{N}'$ into $\mathcal{ALC}$.}
%\label{tab:translation2}
%\end{table}

Checking the consistency of the normative system $\mathcal{N}'$ corresponds to checking the consistency of $\phi(\mathcal{N}')$ w.r.t. the TBox $\mathcal{T}= \lbrace \top \sqsubseteq \exists R.\top\rbrace$, where $\phi(\mathcal{N}')$ is the conjunction of the concepts given in the right column of Table~\ref{tab:translation2}.
We transform $\phi(\mathcal{N}')$ into DL-clauses, which is the input language of Hyper. We will not give the result of this transformation and refer to \cite{msh07optimizing} for details.
Hyper constructs a hypertableau for the resulting set of DL-clauses. This hypertableau is closed and therefore we can conclude, that the set of DL-clauses is unsatisfiable. This tells us, that the above formalized normative system $\mathcal{N'}$ is inconsistent.

\section{Conclusion}
The goal of this paper was twofold: on one side we wanted to show that deontic logic can be very well used to model various phenomena in human reasoning. And secondly, we wanted to demonstrate that an automated theorem proving system, like Hyper, can be used to decide deontic logic by transforming it into description logic and in DL-clauses. The different performance of humans in different contexts could be explained be combining deontic logic with mental model theory from cognitive science. 

To conclude this paper we want to shortly comment a new area of research, namely the formalization of 'robot ethics'. In multi-agents systems and in robotics one is aiming  to  define  formal rules for  the behavior of  agents. As an  example  consider Asimov's laws, which aim at controlling the relation between robots and humans. In \cite{journals/expert/BringsjordAB06} the authors depict  a small example of two surgery robots which have to deal with ethical codes to perform there work. These codes are given with the help of deontic logic, very much the same as we defined the normative systems in this paper. In \cite{kik} we show how to use Hyper to resolve conflicts in multi-agent systems.

%\newpage
\bibliography{wissen}
\end{document}